\documentclass[10pt,twocolumn,letterpaper]{article}

\usepackage{cvpr}
\usepackage{fancyhdr}
\usepackage{amsmath}
\usepackage{amssymb}
\usepackage{booktabs}
\usepackage{epsfig}
\usepackage{float}
\usepackage{graphicx}
\usepackage{multirow}
\usepackage{times}

\usepackage{tabularx}
\usepackage{cite}

\usepackage[pagebackref=true,breaklinks=true,letterpaper=true,colorlinks,bookmarks=false]{hyperref}
\usepackage[capitalise, noabbrev]{cleveref}

\begin{document}

\title{The Impact of Print-Scanning in Heterogeneous Morph Evaluation Scenarios}

\author{Richard E.~Neddo\\
Clarkson University\\
Potsdam, NY, USA\\
{\tt\small neddore@clarkson.edu}\and
Zander W.~Blasingame\\
Clarkson University\\
Potsdam, NY, USA\\
{\tt\small blasinzw@clarkson.edu}\and
Chen Liu\\
Clarkson University\\
Potsdam, NY, USA\\
{\tt\small cliu@clarkson.edu}
}

\maketitle
\thispagestyle{empty}

\begin{abstract}
   Face morphing attacks pose an increasing threat to face recognition (FR) systems.
   A morphed photo contains biometric information from two different subjects to take advantage of vulnerabilities in FRs.
   These systems are particularly susceptible to attacks when the morphs are subjected to print-scanning to mask the artifacts generated during the morphing process.
   We investigate the impact of print-scanning on morphing attack detection through a series of evaluations on heterogeneous morphing attack scenarios.
   Our experiments show that we can increase the Mated Morph Presentation Match Rate (MMPMR) by up to 8.48\%.
   Furthermore, when a Single-image Morphing Attack Detection (S-MAD) algorithm is not trained to detect print-scanned morphs the Morphing Attack Classification Error Rate (MACER) can increase by up to 96.12\%, indicating significant vulnerability.
\end{abstract}

\section{Introduction}
\label{sec: Introduction}
Face recognition (FR) systems have become one of the most widely used biometric modalities, ranging from security to identification applications in government offices, law enforcement, and visa management. These systems are highly effective in preventing unauthorized access while maintaining low false rejection and acceptance rates, placing them among the best methods to reduce security vulnerabilities~\cite{vgg19-mad, on-vuln, morphed_first}. However, these systems still have susceptibilities, notably in the presence of face-morphing attacks. Face morphing attacks aim to exploit the intrinsic nature of FR classifiers that map biometric templates to a singular identity in a one-to-one map. To achieve this, an attacker creates a single morphed face image incorporating the biometric traits and facial landmarks of two different identities.
The morphed image can cause an FR system to incorrectly register a false accept with \textit{both} identities~\cite{Blasingame2021LeveragingAL, sebastian_gan_threaten, sebastian_on_detection_of_ma_gan, morphed_first, Ferrara2016}. 

\begin{figure}[t]
    \centering
    \caption{Artifacts on Morphed Image after Print-Scanning. Calculated and displayed by subtracting the digital image from the print-scanned image}
    \label{fig: Artifacts on Morphed Image after Print and Scan}
    \begin{subfigure}[t]{0.32\linewidth}%
        \includegraphics[width=\textwidth]{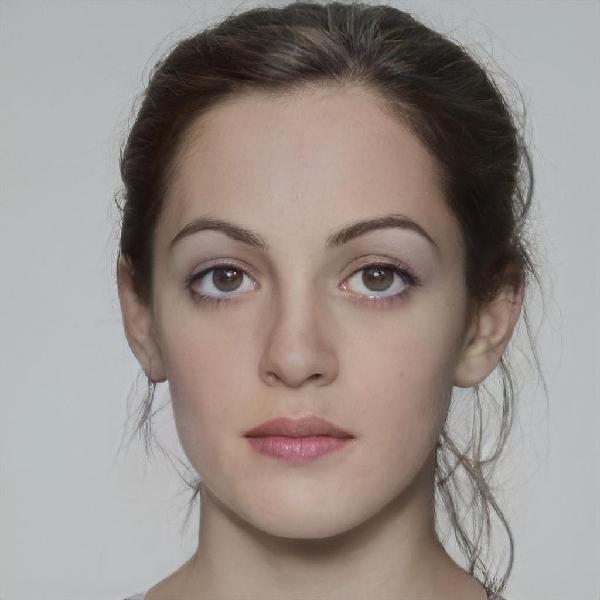}
        \caption{Digitally Morph Image}
        \label{fig: Digitally Morph Image}
    \end{subfigure}
    \begin{subfigure}[t]{0.32\linewidth}%
        \includegraphics[width=\textwidth]{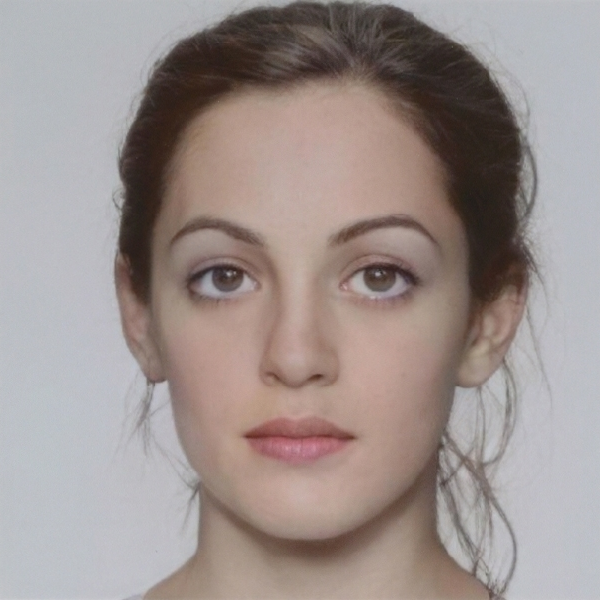}
        \caption{Print-scanned Morphed Image}
        \label{fig: Print-scanned Morphed Image}
    \end{subfigure}
    \begin{subfigure}[t]{0.32\linewidth}%
         \includegraphics[width=\textwidth]{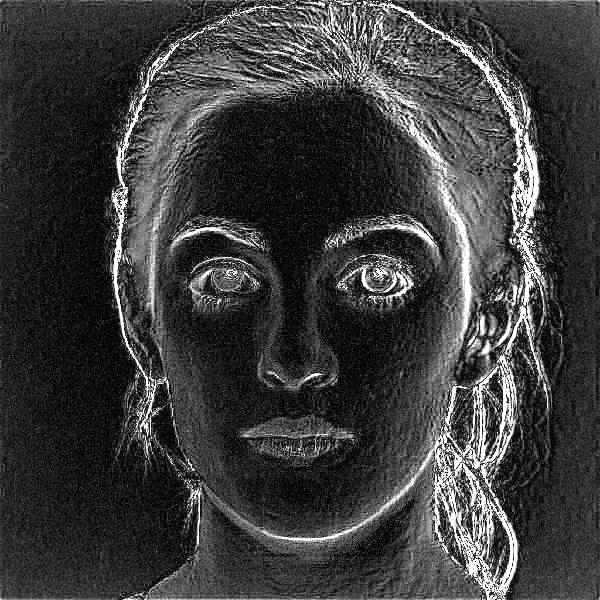}
         \caption{Print-scan Artifacts Amplified}
         \label{fig: Print-scan Artifacts Amplified}
    \end{subfigure}
\end{figure}
One notable concern is the use of morphed images in e-passports and machine-readable travel documents (MRTD)~\cite{Scherhag2019, De_Morphing2019, Hamza2022}. In cases where countries also utilize e-passports for not only renewal but also the issuance of documents, this vulnerability is only amplified~\cite{dim_paper}.
A person who is unauthorized or blacklisted may still be able to get access to restricted systems, areas, or travel if a morphed image is used in an e-passport or MRTD. 
A variety of morphing attacks have been proposed from landmark-based attacks~\cite{can_gan_beat_landmark}, Generative Adversarial Network (GAN) based attacks~\cite{morgan, mipgan}, and diffusion-based attacks~\cite{blasingame2024fastdim,dim_paper,greedy_dim} which pose a grave threat to FR systems.

\begin{figure*}[ht]
    \centering
    \includegraphics[width=0.98\textwidth]{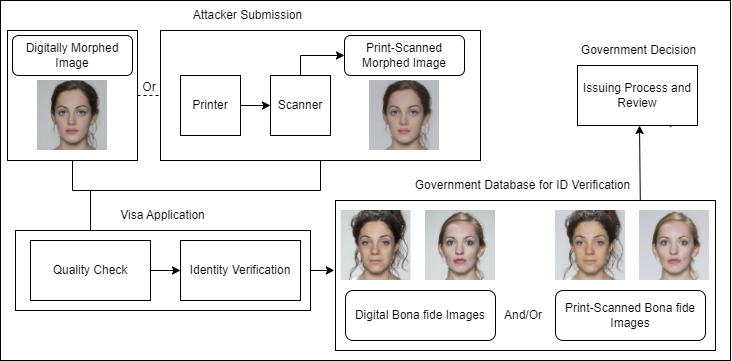}
    \caption{Heterogeneous morph attack pipeline in a simulated real-world scenario}
    \label{fig: Heterogeneous Morph Attack Pipeline}
\end{figure*}

The introduction of print-scan processes only further increases this vulnerability by potentially masking artifacts from the morphing process.
When a morphed image is print-scanned, the physical manipulation can obscure and mask digital artifacts, further complicating detection~\cite{Ferrara_2021, nist-frvt-morph}. 
Physical manipulation of digitally morphed images can remove traces from the morphing algorithm and create a ``new'' image. This attack is particularly effective on FR systems that haven't been trained to incorporate the print-scan style attack since the system cannot classify the unique artifacts generated during printing and scanning. The reason for this is when an image is printed, a unique type of artifact is introduced that masks and destroys the artifacts generated during the digital morphing process. Printing generates marks from where the ink was applied to the paper from rollers and characteristics including the absorption of the ink, the texture of the paper, and how the paper was handled both before and after printing, all causing the original image to be altered in small ways. Scanning will also leave traceable artifacts. The light from the scanner reflects off the page and can be seen in darker areas, like in the pupils, if done incorrectly. The scanner also has the potential to leave digital signatures if various post-processing effects are applied and saved to the image during scanning. The morphed image before print-scanning is shown in~\cref{fig: Digitally Morph Image} and after being print-scanned in~\cref{fig: Print-scanned Morphed Image}. The borders and the noise seen in in~\cref{fig: Print-scan Artifacts Amplified} are the difference between the digital and the print-scanned morphs amplified for easier viewing. The artifacts were introduced during print-scanning when the image was processed and re-digitized. This calculation can be performed by taking the difference of the two morphed images.

To combat the increasing risks associated with face morphing attacks, research and development of Morphing Attack Detection (MAD) algorithms has been conducted to classify if images are either bona fide or morphed. 
The importance of efficient classification stems from the need to identify and mitigate such vulnerabilities discussed above. 
MAD algorithms can be classified into two broad types: Single-image MAD (S-MAD) and Differential MAD (D-MAD)~\cite{mipgan}. 
S-MAD algorithms rely on extensive training to properly identify inconsistencies or irregularities that are characteristic of face morphing, whereas D-MADs benefit from having access to a known bona fide to use for reference increasing the accuracy of detection~\cite{Scherhag2019}.

\noindent\textbf{Contributions.}
While print-scanned morphs have been discussed in previous works~\cite{vgg19-mad, mipgan, nist-frvt-morph, mad-data-eval}, there has been limited research on cross-testing digital and print-scanned morphs with bona fides from both print-scanned and digital sources. This gap in research leaves uncertainties about how print-scanned elements affect FR systems and MAD algorithms in a controlled environment. 
To address this gap, we propose a comprehensive set of experiments designed to quantify and observe the implications of injecting a print-scanned photo into the detection process. By creating a series of heterogeneous attack scenarios, we can systematically evaluate potential vulnerabilities in FRs and MADs when faced with a mix of print-scanned and digital images. This work seeks to fill a gap in existing research and allow for more effective morph detection.

\section{Related works}
While morphing algorithms and FR detection have received strong attention in research, little work has been done to investigate and detect the threat of print-scanned face morphing attacks in a mixed morph attack scenario. Most work has focused on simulated print-scanning procedures or strictly print-scanning morphed photos for analysis.

\noindent\textbf{Print-Scan Simulation}

Matteo~\etal~\cite{Ferrara_2021} focuses on addressing the security threat posed by print-scanned face morphing in electronic identity documents. It highlights the challenges of detecting morphed images in cross-database testing and dealing with print-scanned images. The study introduces novel approaches to train Deep Neural Networks (DNNs) for morphing detection, including generating simulated printed-scanned images, data augmentation strategies, and pre-training on large face recognition datasets. 

A study pixel-to-pixel morphing algorithms was done by Tapia~\etal~\cite{BuschSPS} to create simulated print-scanned images to improve upon datasets consisting of mostly digital photos. The simulated data is compared against images that underwent a manual print-scan process to evaluate the effectiveness of their pixel-to-pixel algorithm.

\noindent\textbf{Print-Scanning}

Ngan~\etal~\cite{nist-frvt-morph} addresses the need for datasets to train algorithms and FR systems for detection is of high importance. In that work under 7,000 images have undergone print-scanning.
The analysis, performed using a subset of Visa-like images, underscores a critical concern: algorithms vary widely in their ability to discern between legitimate and morphed images once they have undergone the print-scan cycle. Algorithms showing low morph miss rates but very high false detection rates indicate these algorithms might classify most scanned photos as morphs even when they are not. Conversely, some single-image morph detectors exhibit low false detection rates but high morph miss rates, suggesting the potential reduction or elimination of morphing artifacts during the print-scan process.

In their work, Raja~\etal\cite{mad-data-eval} highlights the significance of the print-scan process as it reflects a real-world scenario in issuing identity documents. Printing and scanning a morphed image replicates the procedure seen in getting a passport photo or identity card photo taken. Due to the nature of print-scanning, the process can degrade image quality by introducing artifacts, changing image resolution, and altering color and contrast which can cause changes in the way FR systems perceive and quantify data. These changes could potentially impact the effectiveness of MAD algorithms by obscuring the artifacts that are in a morphed image.

An in-depth analysis performed by Zhang~\etal\cite{mipgan} leads to their proposal of a novel approach to generating face morphs that are harder for FR systems to detect. MIPGAN (Morphing through Identity Prior driven GAN) is meant to minimize the introduction of artifacts and leverage a different loss function to increase quality. They also performed a series of experiments under different conditions including digital, print-scanned, and compressed images after print-scanning. These experiments were done to investigate the impact that various processing methods have on detection via MADs. Furthermore, training S-MAD mechanisms with MIPGAN-I generated samples resulted in excellent detection performance on MIPGAN-II samples, especially for print-scan and print-scan compressed data. Human analysis was also conducted leading to the conclusion that print-scanned images were generally harder for both groups to detect compared to digital images.

\noindent\textbf{S-MAD}

There have been several works that focus on S-MAD performance.  Raghavendra~\etal\cite{vgg19-mad} investigates the impact of digital morphs and morphs generated from varied print-scanning methods and D-CNNs to train S-MADs for detection in different scenarios. Past works have also shown that when an S-MAD is trained on a specific morphing algorithm or data type the S-MAD will show increased detection on data with similar characteristics~\cite{blasingame2024fastdim, dim_paper}. A generalized S-MAD algorithm utilizing Vision Transformer (ViT) architecture was proposed by Zhang~\etal\cite{Zhang_2024_CVPR} to detect a wide range of morphing traces for inter and intra-dataset studies. Their work indicates effective performance with increased detection rates among all probed morph presentations.

\section{Heterogeneous Morph Evaluation Scenarios}
In this work ``heterogeneous'' defines the varied nature of datasets that consist of digital and print-scanned images, both bona fides and morphed for evaluation purposes.
These images are tested against each other using different FR systems and an S-MAD across a series of attack scenarios, as shown in~\cref{Attack Scenarios}.
The difference in the attack scenarios stems from whether the bona fide or morphed images underwent a print-scan process.
\cref{Attack Scenarios} provides a brief overview of the scenarios which we enumerate in greater detail below.

\begin{table}[h]
\centering
\caption{Attack scenarios to evaluate impact of heterogeneous data}
\footnotesize
\begin{tabularx}{\linewidth}{@{\extracolsep{\fill}}lll}
\toprule
\textbf{Configuration}  & \textbf{Morph}  &	\textbf{Bona Fide} \\
\midrule
D-D		                & Digital	      &	Digital        \\
D-PS			        & Digital	      &	Print-Scanned  \\
PS-D			        & Print-Scanned   &	Digital        \\
PS-PS                   & Print-Scanned	  &	Print-Scanned  \\
\bottomrule
\end{tabularx}
\label{Attack Scenarios}
\end{table}

\noindent\textbf{Scenario 1: Digital vs Digital (D-D).}
The digital-style attack represents an attacker submitting a morphed photo that has not been print-scanned and is submitted for use. The digitally morphed image is then compared to a bona fide image that has not undergone any form of physical or digital manipulation for verification. This is the baseline scenario used in face morphing papers to determine morph efficacy.

\noindent\textbf{Scenario 2: Digital vs Print-scan (D-PS).}
The digital vs print-scan style attack involves an attacker submitting a morphed photo that has not been print-scanned and is submitted for use. The digitally morphed image is then compared to a bona fide image that has been printed and scanned into a database. This attack scenario is most apt to occur when filing for an ID through a web-based application or service but is deterministic on the age or source of the bona fide images accessible for verification. Older systems typically have more print-scanned photos stored.

\noindent\textbf{Scenario 3: Print-scan vs Digital (PS-D).}
The print-scan vs digital style attack involves an attacker submitting a morphed photo that has been print-scanned and is submitted for use. This image is then compared to a bona fide image that has not been manipulated. When filing for a passport or visa with a paper application, a print-scanned photo is used as the media submission. Modern databases contain both print-scanned photos as well as digital photos. When a digital photo is used to verify the print-scanned image, this scenario occurs.

\noindent\textbf{Scenario 4: Print-scan vs Print-scan (PS-PS).}
The print-scan style attack represents an attacker submitting a morphed photo that has been print-scanned and is submitted for use. This image is then compared to a bona fide image that has been print-scanned into a database and is one of the identities in the morph. Like Scenario 3, if instead of a digital photo, one that has been print-scanned is chosen, this scenario will take place.

\noindent\textbf{Relevance.}
With these configurations, the varying artifacts from each representation can be cross-tested to view the impact on FR systems' ability to detect a morphed image in different circumstances. Scenarios D-PS, PS-D, and PS-PS are especially significant in testing the effectiveness of an FR system or MAD. These scenarios illustrate how an attack might appear if executed outside of an experimental environment. When submitting a photo for government identification, it will be compared to photos that are in a government database. Their data contains images that come from re-digitizing physical media or directly from a digital source. In such instances, it is crucial to assess how a morphed image compares to images derived from different mediums or sources. Likewise, It is important to consider all possible attacks since Scenario D-D is the most commonly referenced attack evaluated and analyzed in face morphing research\cite{greedy_dim,blasingame2024fastdim, Ferrara_2021, Ferrara2016, Scherhag2019}, which serves as an important baseline for comparison purposes. Similarly, Scenario PS-PS illustrates how physical manipulation of media impacts performance against MADs and FRs.

\section{Print-Scan Methodology}
Before being print-scanned, the images to be print-scanned are resized and saved as PNG files. Images are digitally arranged on an $8.5 \times12 $ inch blank PNG. JavaScript scripts are used to send the pages to Adobe Photoshop for print management to maintain ICC profiles.
All print-scanned images are set at a $600\times600$ resolution with a pixel-per-inch value of 300 to replicate a passport photo of size two inches by 2 inches while also maintaining their original aspect ratio.

After printing and scanning, the images go through a series of post-processing scripts that isolate, crop, rename, and resize images to reflect the original digital image for testing, evaluation, and quality control purposes. Photos are saved as a Portable Network Graphics (PNG) file to retain high-quality images. Doing so will prevent the introduction of compression artifacts during post-processing, which could impact FR performance and detection. It should be noted that JPEG compression may be used but is outside of the scope of this work as introducing compression artifacts into images before evaluating their effectiveness against an FR system has been noted to impact performance~\cite{Scherhag2022}. 

\begin{figure*}[ht]
     \centering
     \caption{Image array displaying the importance of setting management during print-scanning to generate high-quality images}
    \label{fig: Importance of Quality Assurance}
     \begin{subfigure}[b]{0.195\textwidth}
         \centering
         \includegraphics[width=0.98\textwidth]{Images/001_002_Digital.jpg}
         \caption{Base digital morphed image}
         \label{fig: Original Image}
     \end{subfigure}
          \begin{subfigure}[b]{0.195\textwidth}
         \centering
         \includegraphics[width=0.98\textwidth]{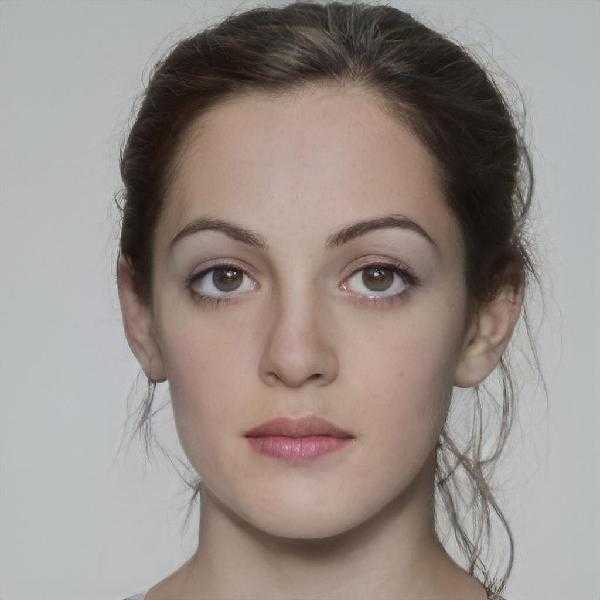}
         \caption{Digital morphed image with compression}
         \label{fig: Compressed Digital Morph}
     \end{subfigure}
     \begin{subfigure}[b]{0.195\textwidth}
         \centering
         \includegraphics[width=0.98\textwidth]{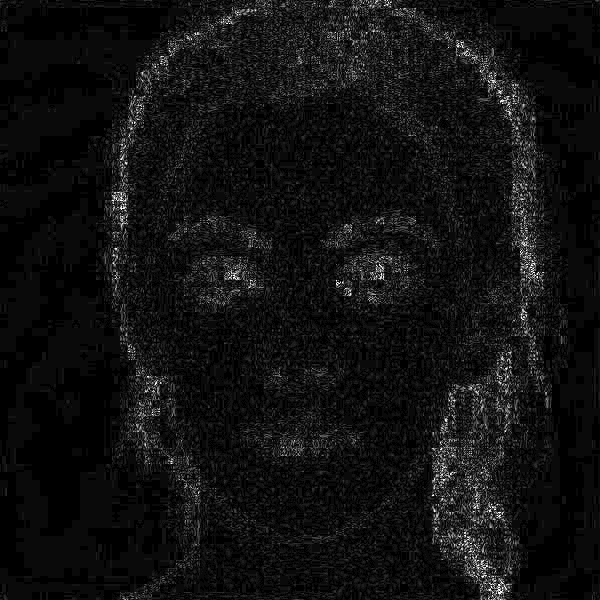}
         \caption{Artifacts on digital morph after compression}
         \label{fig: Compression Artifacts}
     \end{subfigure}
     \begin{subfigure}[b]{0.195\textwidth}
         \centering
         \includegraphics[width=0.98\textwidth]{Images/001_002a.png}
         \caption{Print-scanned morph with proper ICC profile}
         \label{fig: Proper ICC Profile}
     \end{subfigure}
     \begin{subfigure}[b]{0.195\textwidth}
         \centering
         \includegraphics[width=0.98\textwidth]{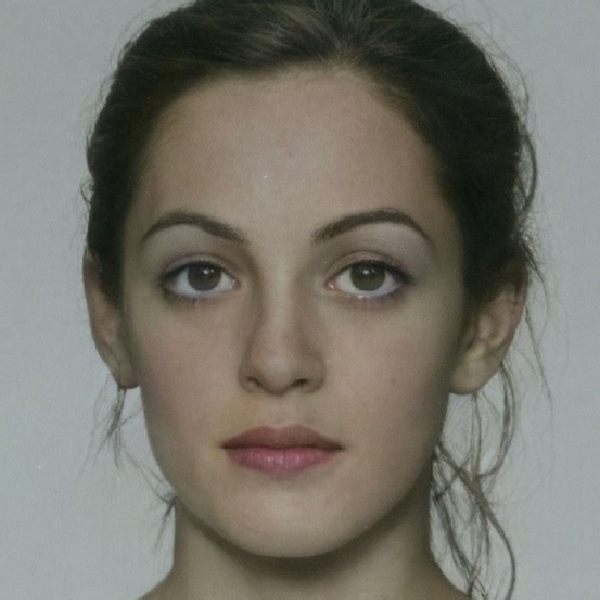}
         \caption{Print-scanned morph with improper ICC profile}
         \label{fig: Improper ICC Profile}
     \end{subfigure}
\end{figure*}

Printed images can develop flaws when International Color Consortium (ICC) profiles are improperly managed~\cite{ICCProfileSpecification43}. ICC profiles manage the exposure, saturation, and hue during printing so when the incorrect profile is used, irregularities will occur, making the resulting print inaccurate. Photo printing paper has a linked ICC profile that must be selected before printing so the printing software can adjust settings accordingly. The implications of improper ICC profile management and compression artifacts can be observed in Figure~\ref{fig: Importance of Quality Assurance}. Photo printing paper was also stored and handled with special care. Cotton gloves were always worn, and photo-printing paper was stored in a plastic sleeve before printing. These precautions were vital in keeping the photo paper from developing a warped surface or getting body oil on the print surface. The scan bed was also wiped with a microfiber cloth to remove dust and other particulates that settled on the surface between scans to avoid adding flaws to the image.

\subsection{Equipment and Settings}
A Canon Pixma Pro 100 Printer and Epson 850v Pro Scanner were used to print and scan all images evaluated in this paper. Printing was handled using JavaScript and Extend-Script debugging scripts to interface with Adobe Photoshop for ICC Color Profile management to ensure proper image quality retention. 
Default scanner software settings include presets and tools to improve image quality by increasing contrast or an image mask to the scanned image before saving occurs. The image quality improvements are fundamentally the same as altering an image in Photoshop, so image enhancement settings must be turned off before saving the image to avoid unintentionally reintroducing artifacts into the scanned images. Before scanning can occur, a sheet of images must be placed on the clean scanner bed and used to calibrate the scanner using the preview feature. Doing so allows settings to be adjusted and verified before scanning a printed dataset. 

\section{Studied Morphing Attacks}
In this study, we explore three different kinds of morphing attacks, representing a broad spectrum of current morphing techniques.

\noindent\textbf{OpenCV.} The na\"ive approach to constructing a face morph is to simply take a pixel-wise average of the two bona fide images; however, this can cause many artifacts especially if the two images are unaligned.
To remedy this, the bona fide images can be aligned using face landmarks ahead of time before applying the pixel-wise average.
These kinds of morphing attacks are known as landmark-based attacks~\cite{Blasingame2021LeveragingAL,sebastian_gan_threaten, can_gan_beat_landmark,on-vuln}. 
OpenCV landmark-based morphs were created by Sarkar~\etal~\cite{sebastian_gan_threaten} using an adaptation of the open-source OpenCV library with the 68-point annotator from the Dlib library~\cite{dlib}.
The face landmarks from the two bona fide images are used to form Delaunay triangles which are then warped and blended to create the final morphed image.

\noindent\textbf{StyleGAN2.}
Further research on face morphing has made use of powerful generative models to perform representation-based morphing attacks~\cite{morgan}.
One type of these generative models is Generative Adversarial Network (GAN), a state-of-the-art single-step image generation model that aims to learn the data distribution by training in an adversarial manner.
The generator network learns a mapping from the latent space to image space, allowing samples to be drawn from this latent space to sample the data space.
For face morphing attacks an encoding strategy is deployed so that for a given image the latent code that represents it is found.
The bona fide images used in the creation of the morphing attack are encoded into their latent representations.
The latents are then morphed by averaging the latent vectors, resulting in the morphed latent.
This morphed latent is then fed to the generator network constructing the morphed face.
Sarkar~\etal~\cite{sebastian_gan_threaten} create face morphs using the very representational powerful StyleGAN2~\cite{stylegan2} architecture.

\noindent\textbf{DiM.}
Blasingame~\etal~\cite{dim_paper} propose \textbf{Di}ffusion \textbf{M}orphs (DiM) which use another kind of state-of-the-art generative model known as diffusion models.
Diffusion models outperform GAN models on image synthesis; however, this comes at the cost of greater computational complexity during inference~\cite{diff_beat_gan}.
A diffusion process is described by an It\^o Stochastic Differential Equation (SDE) which slowly adds white noise to the original image, eventually degrading into pure white noise.
Diffusion models learn how to reverse this SDE, enabling the sampling of images by starting with white noise and iteratively removing the noise over many steps until the denoised image is left.
DiM uses a Diffusion Autoencoder~\cite{diffae} which is also conditioned on additional latent representation of the original image and provides an encoding strategy for mapping the original image into noise.
Both the latent conditional and noise for the morphed images combined to create the morphed representation.
The morphed conditional is constructed by averaging the conditionals of the two bona fide images.
Conversely, the encoded noise of both images is blended using spherical interpolation with a factor of 0.5~\cite{dim_paper}.
The Diffusion Autoencoder then uses the morphed conditional and noise to create the morphed image.
In particular, we use the DiM-C variant which includes an additional pre-processing step before encoding the bona fide images.

\section{Experimental Setup}
To evaluate the effectiveness of print-scanning, morphing attacks were performed using three datasets against three different FR systems and an S-MAD. All computation and evaluation was performed on two systems: 
One has dual Intel Xenon Silver 4114 CPUs and an NVIDIA Tesla V100 32GB GPU with CUDA version 10.1 and CUDNN version 8.4.
The other system uses a Ryzen 9 5900x CPU and an NVIDIA Geforce RTX 3090TI 24GB GPU with CUDA version 11.3 and CUDNN version 8.4. The proposed morphing attacks, MAD algorithms, and FR systems were implemented in PyTorch. The print management and Photoshop scripts were implemented using extendscript. Other post-processing or landmark visualization scripts were implemented using Python.

\subsection{Face Recognition Systems}
Three well-known FR systems are utilized to evaluate the strength of print-scanned morphing attacks: %
ElasticFace~\cite{elasticface}, Adaface~\cite{adaface}, and ArcFace~\cite{deng2019arcface}.%
These FR systems determine a match by comparing feature vectors and finding images with various features within a user-defined tolerance. The ArcFace was trained on the Glint360K dataset, which consists of 17,091,657 images from 360,232 individuals~\cite{glint360k_paper}. The ElasticFace and AdaFace models are trained on the MS1M-ArcFace dataset. All three FR systems use different pre-processing pipelines to present the image as a passport photo. 
ArcFace, ElasticFace, and Adaface models use resized images. The image is cropped to a resolution of $112\times112$ and then normalized to have values in [-1, 1] while maintaining the original aspect ratio.

\subsection{Datasets}
In this study, the FRGC v2.0~\cite{frgc}, FRLL~\cite{frll}, and FERET~\cite{feret} datasets were used to study the impact of print-scanning on face morphing attacks. These datasets were chosen because they are used in NIST face morphing reports~\cite{nist-frvt-morph} and other face morphing analysis papers~\cite{mad-data-eval, zhang2021mipgan}. Sarkar~\etal~\cite{sebastian_gan_threaten} created the StyleGAN2 and OpenCV morphs for the FRGC, FRLL, and FERET datasets.
We implemented the DiM algorithm from~\cite{dim_paper} to create the DiM-C morphs for the FRGC, FRLL, and FERET datasets.
The morphs, component identity pairs, and alternate bona fide identity images were print-scanned for evaluation. This resulted in 8,142 morphs and 4,653 bona fide images being print-scanned, as shown in~\cref{tab: Dataset Breakdown Investigated}. The bona fide images for each dataset are shared between morphing attacks. This work used the bona fide pairs developed in~\cite{sebastian_gan_threaten} for our FRGC, FERET, and FRLL pairings and was used to create the DiM, OpenCV, and StyleGAN2 morphs.

\begin{table}
    \caption{Breakdown of datasets and morphing algorithms}
    \centering
    \footnotesize
    \begin{tabular}{llrr} 
        \toprule
        \textbf{Dataset} & \textbf{MA} & \textbf{Morphs} & \textbf{Bona fides} \\ 
        \midrule
        \multirow{3}{*}{FRLL}       & OpenCV        & 1,221    & \multirow{3}{*}{204}   \\ 
                                    & StyleGAN2     & 1,221    & \\ 
                                    & DiM-C         & 1,221    & \\
        \midrule
        \multirow{3}{*}{FRGC}       & OpenCV        & 964      & \multirow{3}{*}{3,038} \\
                                    & StyleGAN2     & 964      & \\ 
                                    & DiM-C         & 964      & \\
        \midrule                          
        \multirow{3}{*}{FERET}      & OpenCV        & 529      & \multirow{3}{*}{1411}  \\
                                    & StyleGAN2     & 529      & \\ 
                                    & DiM-C         & 529      & \\
        \midrule
        Total                       &               & 8,142    & 4,653  \\
        \bottomrule
    \end{tabular}
    \label{tab: Dataset Breakdown Investigated}
\end{table}

\begin{table*}[t]
\centering
\caption{MMPMR for all scenarios with FMR = 0.1\%. A higher MMPMR value represents a stronger attack.}
\label{MMPMR}
\footnotesize
\begin{tabular}{ll rrr rrr rrr}
\toprule
&& \multicolumn{3}{c}{\textbf{FRLL}} & \multicolumn{3}{c}{\textbf{FRGC}} & \multicolumn{3}{c}{\textbf{FERET}}\\
\cmidrule(lr){3-5}
\cmidrule(lr){6-8}
\cmidrule(lr){9-11}
\textbf{Morph}   & \textbf{Scenario} &
\textbf{ArcFace} & \textbf{ElasticFace} & \textbf{AdaFace} &
\textbf{ArcFace} & \textbf{ElasticFace} & \textbf{AdaFace} &
\textbf{ArcFace} & \textbf{ElasticFace} & \textbf{AdaFace}\\
\midrule
\multirow{4}{*}{OpenCV}    & D-D     & 99.02          & \textbf{98.69} & \textbf{99.26} & 67.31           & \textbf{50.99}& 53.22          & 89.04          & 75.61          & 81.78    \\
                           & D-PS    & \textbf{99.18} & 97.22          & 99.02          & 68.91           & 47.81         & 53.96          & \textbf{89.97} & \textbf{81.66} & \textbf{83.51}    \\
                           & PS-D    & 98.61          & 96.81          & 97.87          & 55.67           & 43.15         & 45.36          & 86.45          & 78.48          & 81.95    \\
                           & PS-PS   & 98.85          & 94.19          & 99.02          & \textbf{69.89}  & 41.61         & \textbf{55.51} & 88.82          & 78.58          & 77.13    \\
\midrule          
\multirow{4}{*}{StyleGAN2} & D-D     & 5.89           & 3.27           & 6.55           & \textbf{1.38}   & 1.21          & 1.25           & 0.82           & 0.32           & 0.72     \\
                           & D-PS    & 3.44           & \textbf{5.56}  & 4.66           & 0.67            & \textbf{1.28} & \textbf{1.45}  & \textbf{0.82}  & \textbf{0.41}  & \textbf{1.29}\\
                           & PS-D    & 5.32           & 1.31           & \textbf{7.53}  & 1.00            & 1.00          & 0.56           & 0              & 0              & 0        \\
                           & PS-PS   & \textbf{6.63}  & 3.11           & 6.38           & 0.41            & 0.44          & 1.36           & 0              & 0              & 0        \\
\midrule              
\multirow{4}{*}{DiM-C}     & D-D     & 92.88          & 82.00          & 88.22          & 48.70          & \textbf{43.24} & 41.75          & 69.76          & 59.65          & 65.27    \\
                           & D-PS    & 90.10          & \textbf{88.95} & 87.81          & 43.65          & 39.23          & 42.66          & 71.53          & 62.39          & 68.46    \\
                           & PS-D    & 92.39          & 77.09          & \textbf{91.33} & \textbf{49.11} & 37.98          & 35.82          & \textbf{74.03} & 62.21          & 65.08    \\
                           & PS-PS   & \textbf{93.62} & 83.22          & 90.83          & 37.47          & 28.30          & \textbf{44.04} & 66.91          & \textbf{64.20} & \textbf{69.99}    \\
\bottomrule
\end{tabular}
\end{table*}

\subsection{Metrics}
Several representative metrics used in face-morphing research were employed in this study. 
The \textit{Mated Morph Presentation Match Rate} (MMPMR) proposed by Scherhag~\etal~\cite{mmpmr} is a widely used metric in evaluating the performance of a morphing attack~\cite{nist-frvt-morph, morphing_attack_potential}.
The MMPMR is defined as
\begin{equation}
M(\delta) = \frac{1}{M}\sum_{m=1}^{M} \left\{ \left[ \min_{n \in \{1,\ldots,N_m\}} S_m^n \right] > \delta \right\}
\end{equation}
where $\delta$ is the verification threshold, $S_m^n$ is the similarity score of the $n$-th subject of morph $m$, $M$ is the total number of morphed images, and $N_m$ is the total number of subjects contributing to morph $m$ (often $N_m = 2$).
In the scenario in which multiple samples of a single subject are compared to a single morphed image, Scherhag~\etal~\cite{mmpmr} recommend using the ProdAvg-MMPMR given as
\begin{equation}
    M(\delta) = \frac1M \sum_{m=1}^M \left[\prod_{n=1}^{N_m}\bigg(\frac{1}{I_m^n}\sum_{i=1}^{I_m^n}\{S_m^{n,i} > \delta\}\bigg)\right]
\end{equation}
where $I_m^n$ is the number of samples of the subject $n$ compared to morph $m$.
As our evaluation compares multiple samples of the original subjects to a single morphed image, we use the ProdAvg-MMPMR, hereafter referred to as the MMPMR.

Another widely used biometric metric is the \textit{Morphing Attack Classification Error Rate} at a \textit{Bona Fide Presentation Classification Error Rate}, or MACER @ BPCER for short. This metric quantifies the accuracy of the accuracy of morphing attack detection by measuring the rate at which morphing attacks are incorrectly classified as genuine biometric samples (MACER) while maintaining a specified rate at which genuine biometric samples are incorrectly classified as fraudulent (BPCER). Using these two rates, a clear distinction in the performance of a MAD system can be expressed.
The BPCER values chosen are thresholds at 0.1\%, 1\%, and 5\%, respectively. A lower BPCER indicates a more strict classification system whereas a high APCER is a sign of a secure system with few false acceptances. It is important to view the trade-offs across a range of rates as a system that cannot identify bona fide images properly is not of too much use. Likewise, MACER @ BPCER can be used for risk assessment and to evaluate the robustness of an FR system or MAD.
The \textit{Equal Error Rate} (EER) represents the point at which the rate of false acceptances equals the rate of false rejections. This is the threshold at which an unauthorized user is just as likely to get into a system as they are rejected and helps indicate the overall accuracy and reliability of a biometric system. A lower EER signifies a system adapted to rejecting unauthorized users while accepting authorized individuals. 

\section{Results} 
\label{sec: Results}
To assess the impact of print-scanned morphs and bona fide images, we conduct experiments considering attacks that use images from different sources. The proposed experiments use the D-D scenario as a baseline as seen in numerous other works~\cite{mipgan,morph_pipe,dim_paper,greedy_dim,blasingame2024fastdim}. This approach allows for the analysis of print-scanned images through comprehensive vulnerability and detectability studies.

\subsection{Vulnerability Study}
This work aims to introduce a novel addition to existing evaluation frameworks. We propose the importance of accounting for heterogeneous media presentations to determine the robustness of FRs. We propose assessing FR vulnerability using the Mated Morphed Presentation Match Rate MMPMR introduced by Scherhag~\etal~\cite{mmpmr}. 

Table~\ref{MMPMR} presents the MMPMR results with a false match rate (FMR) at 0.1\% for each morphing attack, dataset, and evaluation scenario. It should be noted that DiM-C and OpenCV morphs perform better than their GAN-based counterparts. Likewise, the landmark-based morphing algorithm outperforms the diffusion-based morphing algorithm. The high performance of OpenCV and the poor performance of StyleGAN2 are consistent with prior work and the state of the effectiveness of face morphing research~\cite{sebastian_gan_threaten, Blasingame2021LeveragingAL, blasingame2024fastdim, dim_paper}.

Our proposed approach illustrates the impact of heterogeneous media types on every scenario for all available data. Similar performance trends can be observed across all morphing algorithms and FRS, specifically, attack scenarios that contain at least one print-scanned element. 
Vulnerability can be observed in all three datasets and all three FRs. 67\% of the DiM-C PS-PS morphs perform better than the D-D scenario at an average of 4.26\%. When looking at any DiM-C morph scenario containing a print-scanned element the scenarios perform better 89\% of the time at an average of 5.01\%. The maximum difference is 8.48\%
Similar performance can be observed across the OpenCV scenarios that contain a print-scanned element. 67\% of the morph scenarios perform better than the D-D scenario as a baseline averaging 3.17\%. The maximum difference is MMPMR of 8\% for the OpenCV morphing algorithm.

Overall, the varied results indicate that introducing a print-scanned element into an attack scenario instigates unpredictability in the FRS's ability to verify the test subject, thus creating vulnerability. 

\begin{table*}[t]
    \centering
    \caption{S-MAD Study with training by varying OpenCV Morphs with bona fides on FRGC.}
    \footnotesize
    \begin{tabularx}{\linewidth}{@{\extracolsep{\fill}}llrrrrrrrrrrrr}
    \toprule
     &&\multicolumn{4}{c}{\textbf{Digital}}
     &\multicolumn{4}{c}{\textbf{Digital + Print-Scan}}
     &\multicolumn{4}{c}{\textbf{Print-Scan}}\\
     \cmidrule(lr){3-6}
     \cmidrule(lr){7-10}
     \cmidrule(lr){11-14}
     &&&\multicolumn{3}{c}{\textbf{MACER @ BPCER}}
     &&\multicolumn{3}{c}{\textbf{MACER @ BPCER}}
     &&\multicolumn{3}{c}{\textbf{MACER @ BPCER}}\\
     \cmidrule(lr){4-6}
     \cmidrule(lr){8-10}
     \cmidrule(lr){12-14}
     \textbf{Morphing Attack} & \textbf{Scenario} & 
     \textbf{EER} & \textbf{0.1\%} & \textbf{1.0\%} & \textbf{5.0\%} &
     \textbf{EER} & \textbf{0.1\%} & \textbf{1.0\%} & \textbf{5.0\%} &
     \textbf{EER} & \textbf{0.1\%} & \textbf{1.0\%} & \textbf{5.0\%}\\
    \midrule             
\multirow{4}{*}{OpenCV} & D-D     & 0              & 0              & 0              & 0              &  0             & 0              & 0             & 0             & 4.81             & 71.76          & 26.93          & 4.64           \\
                        & PS-D    & 0.82           & 77.25          & 0.63           & 0.13           &  0             & 0              & 0             & 0             & 0                & 0              & 0              & 0              \\
                        & D-PS    & 0              & 0              & 0              & 0              &  0             & 0              & 0             & 0             & \textbf{11.78}   & \textbf{88.55} & \textbf{61.32} & \textbf{26.66} \\
                        & PS-PS   & \textbf{13.63} & \textbf{96.12} & \textbf{70.7}  & \textbf{39.37} &  0             & 0              & 0             & 0             & 0                & 0              & 0              & 0              \\
    \midrule
\multirow{4}{*}{StyleGAN2} & D-D     & 0.13           & 0.13           & 0.07           & 0              & 0.1            & 0.1            & 0             & 0             &  9.97            & 97.33          & 78.41          & 30.74          \\
                           & PS-D    & 6.65           & 96.51          & 47.56          & 10.14          & 0.23           & 0.49           & 0             & 0             &  0.43            & 7.04           & 0.03           & 0              \\
                           & D-PS    & 1.91           & 68.6           & 5.96           & 0.56           & 0.86           & 7.83           & 0.79          & 0.1           & \textbf{25.61}   & \textbf{99.61} & \textbf{85.39} & \textbf{65.01} \\
                           & PS-PS   & \textbf{31.47} & \textbf{99.74} & \textbf{97.5}  & \textbf{79.66} &  \textbf{2.57} & \textbf{48.85} & \textbf{6.65} & \textbf{1.09} &  2.27            & 48.45          & 8.62           & 0.69           \\
    \midrule
\multirow{4}{*}{DiM-C}  & D-D     & 7.67           & 87.03          & \textbf{55.63} & 13.2           & \textbf{15.14} & \textbf{99.8}  & \textbf{91.67} & \textbf{52.21} & \textbf{39.43} & 99.57          & 96.61          & 87.2           \\
                        & PS-D    & 7.9            & \textbf{92.43} & 44.67          & 14.35          & 1.55           & 46.18          & 2.24           & 0.43           & 2.7            & 67.18          & 5.76           & 0.72           \\
                        & D-PS    & 0.3            & 4.34           & 0              & 0              & 1.25           & 20.67          & 1.58           & 0.26           & 36.87          & \textbf{100}   & \textbf{99.61} & \textbf{92.13} \\
                        & PS-PS   & \textbf{9.97}  & 87.52          & 50.2           & \textbf{23.5}  & 2.9            & 68.89          & 7.27           & 1.15           & 7.27           & 91.47          & 51.58          & 13.43          \\
    \bottomrule
    \end{tabularx}
    \label{tab:det_study}
\end{table*}

\subsection{Detectability Study}
Following the approach taken in~\cite{dim_paper,blasingame2024fastdim,greedy_dim}, we design the detectability study using a pre-trained SE-ResNeXt101-32x4d network developed by NVIDIA to analyze the impact of print-scanning.
The backbone used for training the S-MAD detector is a ResNeXt101-32x4d model~\cite{resneXt} with added Squeeze-and-Excitation layers~\cite{squeezenet} pre-trained on the ImageNet~\cite{imagenet} dataset.
We opt to use $k$-fold stratified cross-validation to ensure that the class balance between morphed and bona fide images is preserved across each fold and use $k = 5$ folds for our experiments.
We replace the last layer of the network with a fully connected layer with two outputs which denote the log probabilities of the bona fide and morphed classes.
For each training dataset, the network is trained for 3 epochs with an exponential learning rate scheduler and differential learning rates to combat any potential overfitting during training.
The fully connected layer has a learning rate of 0.001 which is reduced exponentially to a learning rate of $10^{-7}$ for each layer further away from the fully connected layer.
To further combat overfitting, the cross entropy loss uses label smoothing with a value of 0.15.
We track the Exponential Moving Average (EMA) of the model weights during training and use these weights during inference.
The EMA decay $\beta_{ema}$ rate is scaled, with the batch size $M$ in accordance to
\begin{equation}
    \beta_{ema} = \bigg(\frac12\bigg)^{\frac{M}{1000}}
\end{equation}
in a manner similar to that of the scaling rule used to update the generator in the Alias-Free GAN~\cite{stylegan3} and recent work has shown the benefit of scaling the EMA decay with batch size~\cite{scale_ema}.
In our experiments, we use a batch size $M = 128$, and therefore $\beta_{ema} \approx 0.915$.
The result of this training procedure is an S-MAD algorithm that achieves a minimum of 98\% class-balanced accuracy on each training fold.

To evaluate the performance of morphs from heterogeneous sources, we trained the S-MAD detector on a combination of bona fide images and OpenCV morphs from the FRGC dataset.
We opt to evaluate on the FRGC dataset as the dataset has a large number of probe images per identity and has been used by prior works~\cite{mipgan, Ferrara_2021}.
Likewise, we train the S-MAD detector on OpenCV morphs due to the excellent performance of landmark-based morphs~\cite{mipgan,dim_paper,blasingame2024fastdim}.
To assess S-MADs detection capabilities we designed three training scenarios to evaluate all possible bona fide dataset compositions: Digital, Print-Scan, and Digital + Print-scan. This approach enables us to make increasingly concise conclusions on the impacts of different types of media.

\noindent\textbf{Digital.}
When the S-MAD is trained on digital OpenCV morphs, the detector can detect digital OpenCV and StyleGAN2 morphs but consequently has a high error rate when attempting to detect images that are made with the diffusion algorithm and morphs that have been print-scanned. The missed detection rates of print-scanned morphs in this scenario get as high as 99.74\%.
   
\noindent\textbf{Print-Scan.}
The inverse relationship is also true. Non DiM-C print-scanned images were detected when the S-MAD was trained on print-scanned data but was unable to detect digitally morphed images. This inverse relationship highlights the flaws inherited during training detectors and why morphs that have been print-scanned perform so well against a detector trained on digital data. The print-scanned trained S-MAD missed digital morphs at a rate of 97.33\%

\noindent\textbf{Digital + Print-Scan.}
The vulnerabilities seen in the digitally trained S-MAD and the S-MAD trained on Print-Scanned images can be resolved by training the detector on digital and print-scanned sources as well as different morph types. This is evident in the middle column of Table~\ref{tab:det_study}. The rate of missed detections seen in the previous two trainings was reduced to 48.85 on the high end and the error rate also dropped to less than 2.57. Current works that investigate the impact of printing and scanning do not account for the data composition that can be seen in large government databases leading to vulnerabilities when left omitted in training.

This experiment identifies key areas of vulnerability in S-MAD systems which can be attributed to training data as MADs rely on high-quality and diverse morphs for accurate detection. Having more morph types for training allows the S-MAD to differentiate and detect artifacts present in each algorithm and presentation. The Digital and Print-Scan trained S-MADs fail to accurately detect artifacts present in morphs of the inverse style. Evidence from the Digital + Print-Scan S-MAD supports this as the detector trained on both presentations of morphs was able to detect digital and print-scanned morphs at high rates of accuracy. It should be noted that DiM-C detection is low as this is attributed to the training being done on OpenCV. When a detector is trained on OpenCV, DiM goes undetected visa versa~\cite{greedy_dim, blasingame2024fastdim, dim_paper}.

\section{Conclusion}
By introducing print-scanned elements into the morph generation and evaluation pipeline, the ability to discriminate between a morphed image and a bona fide image becomes more difficult, because print-scanning can mask digital artifacts generated during morphing.
The heterogeneous evaluation scenarios account for print-scan artifact introduction by setting experiments up to cross-test digital and print-scanned elements together for thorough analysis.
The method of print-scanning followed in this work produced 12,795 high-quality print-scanned bona fide and morphed images. 
The vulnerability study details how asymmetric data can introduce uncertainty into FRs, making classifications for verifying ID difficult. 
Furthermore, the detectability study illustrates the importance of training MADs for generalized detection, as omitting data from training introduces vulnerabilities where attacks can go undetected. 

The scope of this work covers three representative morphing algorithms across three datasets. Repeating the steps followed in this work with additional morphing methods would help support the findings made in Section~\ref{sec: Results}. In terms of future work, performing experiments with more FR systems, more S-MAD studies, and using additional classifying metrics like Morphing Attack Potential~\cite{Ferrara_2021} would provide additional information for allowing for better comparisons between data~\cite{blasingame2024fastdim}. 
Another future work would be to expand the range of equipment and paper used to generate print-scanned images. This evaluation reflects only a portion of the possible impacts associated with ICC profiles, ink, printing, and scanning. Likewise, investigating the introduction and removal of biometric landmarks and artifacts embedded in latent color space would allow for deeper examinations into the unseen impacts of print-scanning~\cite{Batskos2023}.

{\small
\bibliographystyle{IEEEtran}
\bibliography{egbib}
}

\newpage
\onecolumn
\appendix

\section{Additional Results}

To further investigate the performance of morphs from heterogeneous sources, we trained the S-MAD detector on a combination of data from the FRGC dataset. We opt to evaluate using the FRGC dataset as the training dataset as there are a large number of probe images per identity and has been used by prior works~\cite{mipgan, Ferrara_2021}.
We train the S-MAD detector on DiM-C morphs and StyleGan2 morphs to investigate the detector performance with varied input data.
To assess the S-MADs detection capabilities we designed three training scenarios to evaluate all possible bona fide dataset compositions: Digital, Print-Scan, and Digital + Print-scan. This approach enables us to make increasingly concise conclusions on the impacts of varying training data for diverse media.

\begin{table}[h]
    \centering
    \caption{S-MAD Study with training by varying DiM-C Morphs with bona fides on FRGC.}
    \footnotesize
    \begin{tabularx}{\linewidth}{@{\extracolsep{\fill}}llrrrrrrrrrrrr}
    \toprule
     &&\multicolumn{4}{c}{\textbf{Digital}}
     &\multicolumn{4}{c}{\textbf{Digital + Print-Scan}}
     &\multicolumn{4}{c}{\textbf{Print-Scan}}\\
     \cmidrule(lr){3-6}
     \cmidrule(lr){7-10}
     \cmidrule(lr){11-14}
     &&&\multicolumn{3}{c}{\textbf{MACER @ BPCER}}
     &&\multicolumn{3}{c}{\textbf{MACER @ BPCER}}
     &&\multicolumn{3}{c}{\textbf{MACER @ BPCER}}\\
     \cmidrule(lr){4-6}
     \cmidrule(lr){8-10}
     \cmidrule(lr){12-14}
     \textbf{Morphing Attack} & \textbf{Scenario} & 
     \textbf{EER} & \textbf{0.1\%} & \textbf{1.0\%} & \textbf{5.0\%} &
     \textbf{EER} & \textbf{0.1\%} & \textbf{1.0\%} & \textbf{5.0\%} &
     \textbf{EER} & \textbf{0.1\%} & \textbf{1.0\%} & \textbf{5.0\%}\\
    \midrule             
\multirow{4}{*}{OpenCV}    & D-D   &  4.08          & 70.9           & 13.03          & 3.39          &  3.59         & 49.7           & 14.94           & 2.47           & 13.69          & 92.2           & 67.94          & 29.13          \\
                           & PS-D  & 25.18          & 97.63          & 87.56          & 65.54         &  0.3          & 1.55           & 0.2             & 0.07           &  0.03          & 0.03           & 0.03           & 0              \\
                           & D-PS  &  1.78          & 39.53          & 2.83           & 0.53          &  5.69         & 82.55          & 39.8            & 6.81           & \textbf{17.12} & \textbf{96.84} & \textbf{80.09} & \textbf{41.31} \\
                           & PS-PS & \textbf{41.51} & \textbf{98.49} & \textbf{93.42} &\textbf{85.94} & \textbf{15.8} & \textbf{92.36} & \textbf{83.11}  & \textbf{47.7}  &  8.29          & 94.31          & 50.63          & 13.66          \\
    \midrule
\multirow{4}{*}{StyleGAN2} & D-D   &  8.72          & 97.2           & 46.38          & 16.66          & 2.17           & 80.94          & 5.92          & 0.36           & 6.22           & 84.69          & 51.48          & 7.27           \\
                           & PS-D  & 17.38          & 98.49          & 84.13          & 57.93          & 0.36           & 0.56           & 0.26          & 0.07           & 0.3            & 0.63           & 0.03           & 0              \\
                           & D-PS  & 10.53          & 91.08          & 60.5           & 27.52          & 7.67           & \textbf{98.12} & 52.01         & 14.02          & \textbf{18.27} & \textbf{99.93} & \textbf{88.78} & \textbf{57.04} \\
                           & PS-PS & \textbf{33.18} & \textbf{99.77} & \textbf{95.06} & \textbf{81.34} & \textbf{11.09} & 94.6           & \textbf{81.5} & \textbf{30.22} & 6.75           & 91.71          & 32.13          & 8.69           \\
    \midrule
\multirow{4}{*}{DiM-C}     & D-D   &  0           & 0              & 0              & 0             &  \textbf{0.07} & \textbf{0.07}     & 0             & 0              & \textbf{11.52} & \textbf{99.08} & \textbf{87.66} & \textbf{33.67} \\
                           & PS-D  &  2.07        & \textbf{69.95} & \textbf{10.43} & 0.33          &  0             & 0                 & 0             & 0              & 0              & 0              & 0              & 0              \\
                           & D-PS  &  0           & 0              & 0              & 0             &  0             & 0                 & 0             & 0              & 1.91           & 38.71          & 4.11           & 0.95           \\
                           & PS-PS & \textbf{2.5} & 65.67          & 8.13           & \textbf{0.92} &  0.03          & 0.03              & 0             & 0              & 0.1            & 0.39           & 0              & 0              \\
    \bottomrule
    \end{tabularx}
    \label{tab:det_study_DiM}
\end{table}

\begin{table}[h]
    \centering
    \caption{S-MAD Study with training by varying StyleGAN2 Morphs with bona fides on FRGC.}
    \footnotesize
    \begin{tabularx}{\linewidth}{@{\extracolsep{\fill}}llrrrrrrrrrrrr}
    \toprule
     &&\multicolumn{4}{c}{\textbf{Digital}}
     &\multicolumn{4}{c}{\textbf{Digital + Print-Scan}}
     &\multicolumn{4}{c}{\textbf{Print-Scan}}\\
     \cmidrule(lr){3-6}
     \cmidrule(lr){7-10}
     \cmidrule(lr){11-14}
     &&&\multicolumn{3}{c}{\textbf{MACER @ BPCER}}
     &&\multicolumn{3}{c}{\textbf{MACER @ BPCER}}
     &&\multicolumn{3}{c}{\textbf{MACER @ BPCER}}\\
     \cmidrule(lr){4-6}
     \cmidrule(lr){8-10}
     \cmidrule(lr){12-14}
     \textbf{Morphing Attack} & \textbf{Scenario} & 
     \textbf{EER} & \textbf{0.1\%} & \textbf{1.0\%} & \textbf{5.0\%} &
     \textbf{EER} & \textbf{0.1\%} & \textbf{1.0\%} & \textbf{5.0\%} &
     \textbf{EER} & \textbf{0.1\%} & \textbf{1.0\%} & \textbf{5.0\%}\\
    \midrule             
\multirow{4}{*}{OpenCV}    & D-D   & 0.3            & 0.92           & 0.07           & 0              & 1.35           & 65.57          & 2.37           & 0.3           & 18.66          & \textbf{99.9}  & 87.23         & 56.35          \\
                           & PS-D  & 2.47           & 59.97          & 8.99           & 1.15           & 0.1            & 0.1            & 0              & 0             & 0.03           & 0.03           & 0             & 0              \\
                           & D-PS  & 0.53           & 0.79           & 0.36           & 0.03           & \textbf{6.42}  & 77.02          & \textbf{43.38} & \textbf{8.95} & \textbf{34.66} & 99.28          & \textbf{92.4} & \textbf{80.91} \\
                           & PS-PS & \textbf{25.12} & \textbf{96.12} & \textbf{84.86} & \textbf{64.85} & 2.9            & \textbf{90.68} & 20.28          & 1.12          & 1.18           & 10.07          & 1.68          & 0.46           \\
    \midrule
\multirow{4}{*}{StyleGAN2} & D-D   & 0              & 0              & 0              & 0              &  0             & 0              & 0             & 0              & 0.43           & 3.55           & 0.2           & 0            \\
                           & PS-D  & 0.3            & 0.72           & 0.03           & 0              &  0             & 0              & 0             & 0              & 0              & 0              & 0             & 0            \\
                           & D-PS  & 0              & 0              & 0              & 0              &  0             & 0              & 0             & 0              & \textbf{0.82}  & \textbf{20.11} & \textbf{0.79} & \textbf{0.1} \\
                           & PS-PS & \textbf{3.23}  & \textbf{38.51} & \textbf{7.41}  & \textbf{1.22}  &  \textbf{0.03} & \textbf{0.03}  & 0             & 0              & 0.1            & 0.2            & 0             & 0            \\
    \midrule
\multirow{4}{*}{DiM-C}     & D-D   & \textbf{18.8} & \textbf{99.37} & \textbf{92.5} & \textbf{59.05} & \textbf{35.94} & \textbf{99.97} & \textbf{97.47} & \textbf{92.96} & \textbf{51.35} & \textbf{100} & \textbf{99.61} & \textbf{97.43} \\
                           & PS-D  & 5.13          & 90.13          & 35.71         & 5.17           & 1.78           & 76.46          & 5.99           & 0.3            & 2.7            & 67.54        & 11.72          & 0.36           \\
                           & D-PS  & 2.01          & 18.6           & 3.92          & 0.86           & 9.35           & 87.72          & 64.52          & 24.88          & 45.75          & 100          & 98.32          & 94.27          \\
                           & PS-PS & 10.17         & 96.94          & 60.5          & 20.64          & 6.42           & 86.41          & 54.67          & 12.87          & 6.12           & 96.64        & 65.96          & 9.12           \\
    \bottomrule
    \end{tabularx}
    \label{tab:det_study_Gan} 
\end{table}

\noindent\textbf{Digital.}
When the S-MAD is trained on digital DiM-C morphs, the detector struggles to detect non-digital DiM-C morphs as well as all StyleGAN2 and OpenCV morphs. The error and MACER rates for print-scanned DiM-C morphs are lower than the rates seen in the other morphing algorithms but are still undesirable. The missed detection rates of print-scanned morphs in this scenario get as high as 99.77\% seen in Table~\ref{tab:det_study_DiM} in PS-PS of StyleGan2.
The S-MAD trained on StyleGan2 shown in Table~\ref{tab:det_study_Gan} has the same issues as the detector trained on DiM-C. The MACER rates are above 96.12\% for DiM-C and OpenCV print-scan morphs, while detection for digital StyleGan2 morphs is high as expected.
   
\noindent\textbf{Print-Scan.}
Print-scanned DiM-C images were detected when the S-MAD was trained on print-scanned DiM-C data but was unable to detect most print-scanned and digitally morphed images in different morphing styles. The print-scanned trained S-MAD, Table~\ref{tab:det_study_DiM}, missed digital morphs at a rate of 99.08\%. Likewise, missed digital detection rates as high as 100\% can be seen in the S-MAD trained on StyleGan2, Table~\ref{tab:det_study_Gan}, with similar poor performance seen in the Digital evaluation above.

\noindent\textbf{Digital + Print-Scan.}
The rate of missed detections seen in the DiM-C experiment was reduced to 0.07 on the high end and the error rate also dropped to less than 0.07. The rates are much worse for StyleGan2 and OpenCV detections due to the artifacts present being much different than those found in StyleGan2 and OpenCV morphs. The missed detection rates seen in the StyleGan2 S-MAD are lower than those seen in both the digital and print-scan scenarios but are insignificant showing poor performance overall in morphs that aren't StyleGan2. These experiments highlight the need for more generalized algorithms trained to detect multiple morphing algorithms as each S-MAD evaluated has shown clear flaws depending on the attack being performed. 

\end{document}